# MechRetro is a chemical-mechanism-driven graph learning framework for interpretable retrosynthesis prediction and pathway planning

Yu Wang[1,2], Chao Pang[1,2], Yuzhe Wang[1,2], Yi Jiang[1,2], Junru Jin[1,2], Sirui Liang[1,2], Quan Zou[3*], and Leyi Wei[1,2*]

[1] School of Software, Shandong University, Jinan 250101, China

[2] Joint SDU-NTU Centre for Artificial Intelligence Research (C-FAIR), Shandong University, Jinan 250101, China

[3] Institute of Fundamental and Frontier Sciences, University of Electronic Science and Technology of China.

*Corresponding author:
Q.Z: zouquan@nclab.net
L.W: weileyi@sdu.edu.cn

**Abstract**

Leveraging artificial intelligence for automatic retrosynthesis speeds up organic pathway planning in digital laboratories. However, existing deep learning approaches are unexplainable, like "black box" with few insights, notably limiting their applications in real retrosynthesis scenarios. Here, we propose MechRetro, a chemical-mechanism-driven graph learning framework for interpretable retrosynthetic prediction and pathway planning, which learns several retrosynthetic actions to simulate a reverse reaction via elaborate self-adaptive joint learning. By integrating chemical knowledge as prior information, we design a novel Graph Transformer architecture to adaptively learn discriminative and chemically meaningful molecule representations, highlighting the strong capacity in molecule feature representation learning. We demonstrate that MechRetro outperforms the state-of-the-art approaches for retrosynthetic prediction with a large margin on large-scale benchmark datasets. Extending MechRetro to the multi-step retrosynthesis analysis, we identify efficient synthetic routes via an interpretable reasoning mechanism, leading to a better understanding in the realm of knowledgeable synthetic chemists. We also showcase that MechRetro discovers a novel pathway for *protokylol*, along with energy scores for uncertainty assessment, broadening the applicability for practical scenarios. Overall, we expect MechRetro to provide meaningful insights for high-throughput automated organic synthesis in drug discovery.

**Introduction**

Retrosynthesis aims to identify a set of appropriate reactants for the efficient synthesis of target molecules, which is indispensable and fundamental in computer-assisted synthetic planning[1-3]. Retrosynthetic analysis is first formalized by Corey[4-6] and solved by the Organic Chemical Simulation of Synthesis (OCSS) program. Later, driven by sizeable experimental reaction data and significantly increased computational capabilities, various machine-learning-based approaches[7], especially deep-learning (DL) models, have been proposed and achieved incremental performance[8]. From the perspective of molecule representations, existing DL-based retrosynthesis approaches can be generally categorized into two classes: (1) Sequence-based retrosynthesis approaches[9-11], and (2) Graph-based retrosynthesis approaches[12,13]. The former utilizes sequential structures to represent molecules and predicts by encoder-decoder-based neural language architecture, while the latter employs graph structures to describe molecules and forecasts through molding the graph changes between products and reactants.

Sequence-based approaches utilize serialized notations (e.g., SMILES[14], SELFIES[15]) to describe molecules and leverage the encoder-decoder framework to translate target products into potential reactants. Liu et al.[16] present a model, namely Seq2Seq, which consists of a bidirectional long short-term memory (LSTM)[17] encoder and decoder for retrosynthetic translation. Later, with the development of neural machine translation models, like Transformer[18], sequenced-based retrosynthetic approaches achieve gradual performance improvement. Karpov et al.[19] first adapt Transformer architecture with modified learning rate schedules and snapshot learning to retrosynthesis. Segler et al.[9] introduce specialized data augmentation techniques on retrosynthetic Transformers, demonstrating the combination of SMILES augmentation leads to better results. Afterward, as the pretraining-finetuning paradigm rises, Irwin et al.[20] propose MolBART with large-scale and self-supervised pre-training, which significantly speeds up the convergence of the retrosynthesis task. However, there exist the following two limitations. First, the direct structural information in molecules is denoted by implicit symbols, leading to more computational costs and information loss. Second, the popular molecular representation approaches are grammatically strict and not guaranteed semantically valid, resulting in frequent invalid syntaxes. To address the limitations, Ucak et al. propose RetroTRAE[21], alleviating the structural information loss by encoding atom environments; Zheng et al.[22] propose SCROP, effectively avoiding illegal outcomes through coupling with a DL-based syntax corrector.

Graph-based approaches usually adopt graph structures to represent molecules and predict graph changes of the target molecule to infer the reactants. The graph changes are most modeled by the two-stage paradigm that contains reaction center prediction (RCP) and synthon completion (SC). Jin et al.[12] propose WLDN, using the Weisfeiler-Lehman isomorphism test[23] for retrosynthetic graph learning. Later, with the development of graph neural networks (GNNs), many GNN-based frameworks emerged and achieved a notable improvement in performance. Shi et al.[24] present the G2G framework, which utilizes R-GCN[25] for RCP and reinforcement learning for SC. Following the same paradigm, Yan et al.[26] and Somnath et al. [13] devise RetroXpert and GraphRetro, respectively. The former applies a GAT[27] variant for RCP and a sequence-based Transformer for SC, while the latter designs two MPNNs[28] for the two stages. Different from the above approaches, Dai et al.[29] propose GLN, a novel method that leverages reaction templates to connect products and reactants. Nevertheless, traditional GNNs excessively focus on local structures of molecules, neglecting the effect of long-distance characteristics (e.g., Van der Waals' force). To solve the problem, Ying et al.[30] propose Graphormer, introducing a shortest-path-based method for multi-scale topological encoding.

Although existing retrosynthesis approaches have achieved significant progress in accelerating the data-driven retrosynthesis prediction, they still suffer from the following intrinsic problems. (1) Sequence-based approaches suffer from the loss of topological information of molecules susceptible to chemical reactions. Meanwhile, graph-based approaches neglect sequential information and long-range characteristics. Both of them are constrained in feature representation learning, limiting further performance improvement. (2) Existing deep-learning-based retrosynthesis approaches face a common problem; that is, the decision-making mechanism of the models remains unclear, which restricts the model's reliability and practical applications. (3) Most existing approaches focus only on the single-step retrosynthesis prediction that enables generating plausible reactants but is perhaps not accessible. Therefore, the multi-step retrosynthesis prediction with pathway planning from products to accessible reactants is much more meaningful for experimental researchers in practical chemical synthesis.

In this work, we propose MechRetro, an innovative deep graph learning framework via integrating chemical prior knowledge for interpretable retrosynthesis prediction and pathway planning. To learn discriminative molecule representations, we propose a novel Graph Transformer architecture that enables sufficiently capturing both topological information and sequential information of molecules. Comparative results show that MechRetro outperforms the state-of-the-art approaches with a large margin on benchmark datasets. Via interpretable analysis, we showcase MechRetro adaptively captures the discriminative sub-structures of molecules concerning different chemical reactions, indicating the vital capacity in molecule feature representation learning. To simulate the real chemical reaction mechanism and guarantee the chemical rules, we model three reaction steps (e.g., bond changes, leaving group matching, and leaving group connection) through a self-adaptive joint-learning strategy, leading to the transparency of the model decision process and improving the model interpretability. We demonstrate that the self-adaptive joint-learning strategy can learn shared knowledge across different chemical reactions, improving the model's performance and robustness. Importantly, by integrating a heuristic tree-search algorithm into MechRetro, we can identify efficient synthetic routes for drug molecules in the multi-step retrosynthesis prediction, demonstrating the strong ability in high-quality pathway planning. Interestingly, case study results on some well-known drugs show that MechRetro discovers and verifies credible synthetic routes that have not been reported in the literature before, highlighting the practicability in real scenarios. We expect MechRetro to provide meaningful insights for high-throughput automated organic synthesis in drug discovery.

## Methods and materials

### The framework of the proposed MechRetro

The overview of our MechRetro is illustrated in **Fig.1a**. As can be seen, our MechRetro contains four major modules: (1) Graph Transformer module, (2) Self-adaptive learning module, (3) Decision-making module, and (4) Prediction & pathway planning module. The workflow of our MechRetro is described as follows.

In the Graph Transformer module (**see Fig.1b**), taking the processed molecule graph data as input, we first propose a multi-sense and multi-scale bond embedding strategy for the bonds of molecules while introducing topological and contrastive atom embedding strategies for the atoms of molecules, to capture chemically important information. Afterward, the model learns hidden molecular representations from two embedding entries via a multi-head attention mechanism. Next, in the Self-adaptive learning module, the resulting hidden molecular representations are parallelly passed into three deep neural decision units: reaction center predictor (RCP), leaving group matcher (LGM) and leaving group connecter (LGC), which are trained simultaneously by an elaborate self-adaptive learning strategy to maintain the shared knowledge. Among the decision units, RCP identifies the changes of bonds and atoms' hydrogen count (see **Fig.1c**); LGM matches the leaving groups (see section **Reaction center and leaving group** for details) from the collected database for products (see **Fig.1d**); LGC connects the leaving groups and fragments from the product (see **Fig.1e**). In the Decision-making module, the product is transformed into reactants after a decision process consisting of five retrosynthetic actions (see **Fig.1f**) and energy scores for uncertainty assessment, which simulates a reversed chemical mechanism process. In the last module, based on the single-step predictions, a heuristic tree-search algorithm is integrated to discover efficient synthetic routes with transparent decision-making processes while guaranteeing the accessibility of start reactants. The details of the four modules are described as follows.

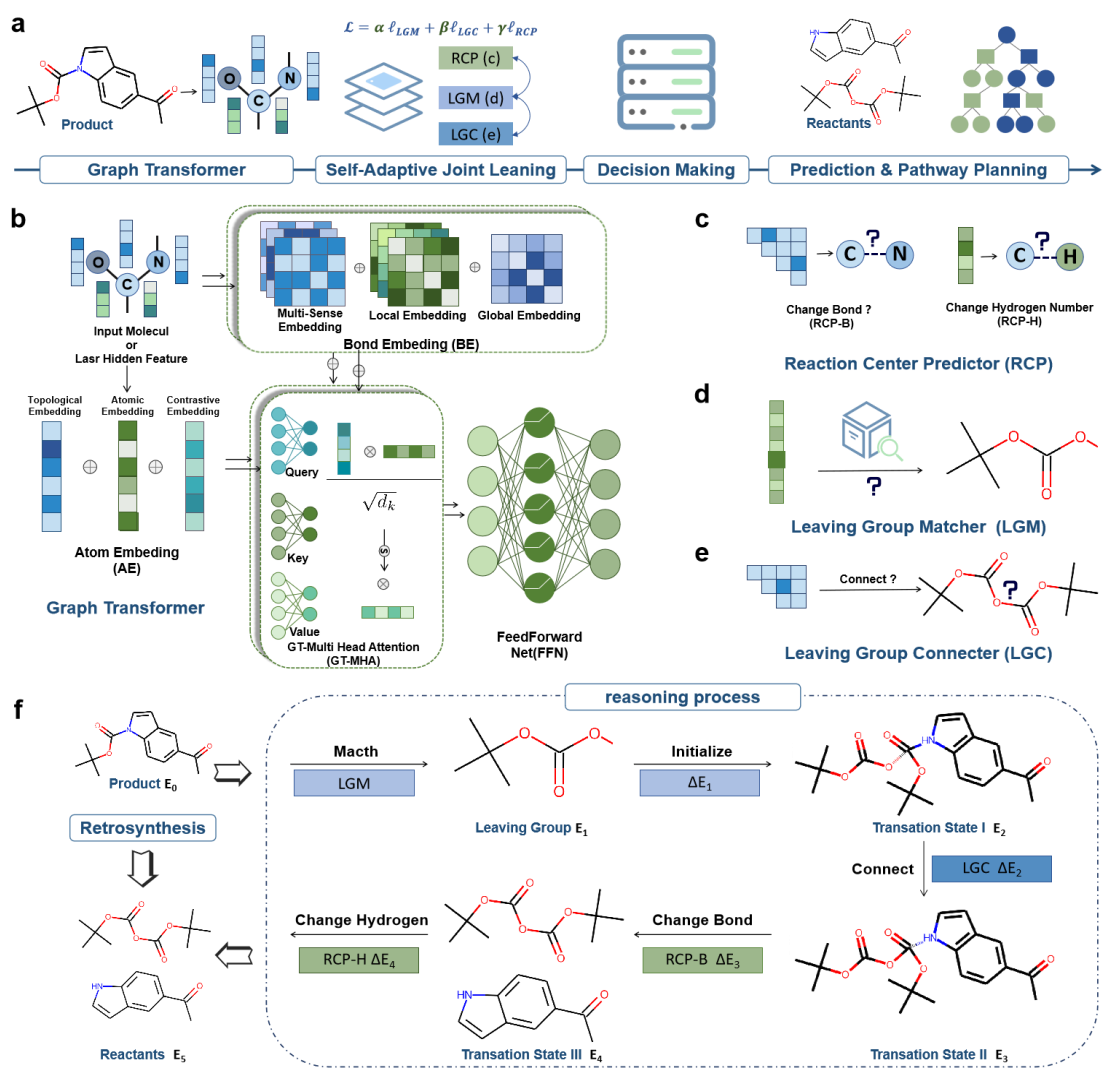

**Fig.1. Overview of MechRetro. a.** MechRetro contains four major modules: (1) Graph Transformer, (2) Self-adaptive learning, (3) Decision making, and (4) Prediction & pathway planning. **b.** MechRetro introduces (1) multi-sense and multi-scale bond embeddings and (2) topological and contrastive atom embeddings for more chemically important information. **c.** RCP recognizes bond changes between product and reactants and hydrogen changes for each atom. **d.** LGM picks out an appropriate leaving group from a collected leaving group set. **e.** LGC identifies which atom in the reaction center should connect to the gate atom of the predicted leaving group by calculating a connection matrix between the original product and the predicted leaving group. **f.** The product is predicted to the reactants via an interpretable reasoning process in our retrosynthesis model. The reasoning process can provide more insights into the retrosynthesis prediction in real scenarios.

## Problem definition

For the convenience of description and discussion., we here briefly introduce the problem definition.

***Graph-based retrosynthesis prediction.*** Generally, a chemical reaction can be denoted as $\left(\{\mathcal{G}_{p;i}\}_{i=1}^{Np}, \{\mathcal{G}_{r;j}\}_{j=1}^{Nr}\right)$, where $\mathcal{G}_p$ represents product graph and $\mathcal{G}_r$ is a reactant graph. And a reaction inference model can be $f_\theta\left(\{\mathcal{G}_{p;i}\}_{i=1}^{Np}\big|\{\mathcal{G}_{r;j}\}_{j=1}^{Nr}\right)$, where $\theta$ is a group of learnable parameters. However, retrosynthesis prediction is not as simple as an inverse model $f_\gamma\left(\{\mathcal{G}_{r;j}\}_{j=1}^{Nr}\big|\{\mathcal{G}_{p;i}\}_{i=1}^{Np}\right)$ for forwarding inference. In practical analysis, the input only includes the main product, which neglects any other by-products as prior knowledge. So, the actual model for retrosynthesis is $f_\gamma\left(\{\mathcal{G}_{r;j}\}_{j=1}^{Nr}, \{\mathcal{G}_{p;i}\}_{i=1}^{Np}/\{\mathcal{G}_{p;m}\}\big|\mathcal{G}_{p;m}\right)$, where $\mathcal{G}_{p;m}$ denotes the main product. It is more complex than the forward prediction, which is why the performance of the forward model is generally better than that of retrosynthesis prediction.

***Reaction center and leaving group.*** In this work, the reaction center is defined as a subgraph $\mathcal{G}_c = \left\{(v,u), e_{uv}|e_{uv} \in \mathcal{G}_p, e_{uv} \notin \mathcal{G}_{r;j}, \forall \mathcal{G}_{r;j} \in \{\mathcal{G}_{r;i}\}_{i=1}^{Nr}\right\}$, where $(u,v)$ denotes an atom pair and $e_{uv}$ is their edge. Also, synthons are a set of graphs modified by-products according to reaction centers. These synthons usually are not chemically valid and can be later converted to final reactants by attaching leaving group. Herein, leaving group $\mathcal{G}_l$ is defined as a subgraph from reactants, which contains reactants atoms and edges that don't occur in the main product. Namely, leaving group records information from by-products $\{\mathcal{G}_{p;i}\}_{i=1}^{Np}/\{\mathcal{G}_{p;m}\}$.

***Transformer from a GNN perspective.*** In the data flow of Transformer, a sentence can be seen as a fully connected graph with the semantic edge, where word tokens are processed as nodes. In this view, MHA can be factorized as follows:

$$m_a^{(l)} := \sum_{w' \in \mathcal{N}_w} \tilde{a}_{w'w} S_{w'}^{(l-1)} W_V, \qquad m_w^{(l)} := \tilde{a}_{ww} S_w^{(l-1)} W_V, \tag{1}$$

$$s.t. \tilde{a}_{w'w} = SOFTMAX\left(\frac{S_{w'}^{(l-1)} W_Q \left(S_w^{(l-1)} W_K\right)^T}{\sqrt{d_k}}\right), \tag{2}$$

$$S''^{(l)}_w = COMBINE^{(l)}\left(m_a^{(l)}, m_w^{(l)}\right) := CONCAT\left(\left\{m_a^{(l;k)} + m_w^{(l;k)}\right\}_{k=1}^{n_{head}}\right) W^O, \tag{3}$$

where $\mathcal{N}_w = S/\{w\}$ is the semantic neighbor set of $w$. Then, **Supplementary Eq. (9)** acts like a non-linear propagation layer in a typical GNN, which transforms $S''^{(l)}_w$ to final $S_w^{(l)}$. Note that the normalized attention matrix $\tilde{A}_s$ composed of $\tilde{a}_{w'w}$ of each word token pair $(w', w)$ can be considered as a group of adaptive parameters which describe the dynamic distribution of semantic edges. However, the standard Transformer cannot handle edges on a topological space. As a variant of Transformer, our MechRetro solve this edge embedding problem.

**Graph Transformer module**

*Atomic and topological embedding.* Given a reaction data $S_r$ represented by SMILES, the related molecule graph of the product $\mathcal{G}_p = (\mathcal{V}, A)$ is constructed, where atoms are viewed as a set of nodes $\mathcal{V}$ with size $N$. To fit such graph data into the Transformer variant, a naive approach considers atoms as word tokens, which are then a reference to build a fully connected graph of the semantic domain. In MechRetro, we adopt this simple method but employ a topological embedding that measures a node's importance in the space domain's graph instead of using conventional position encoding, which destroys original permutation invariance. In detail, we use degree counts of the node to describe the above importance. Therefore, the initial node feature $h_v^{(0)}$ of atom $v$ can be calculated as follows:

$$h_v^{(0)} = \sum_{i=1}^{K_a} \phi_{x_i}(x_i) + \phi_d(\deg(v)), \qquad (4)$$

where $\phi(\cdot)$ with different subscripts denote different embedding functions, $X_i$ is the $i$-th atomic feature (e.g., atomic number, formal charge, etc.), $deg(v) \in \mathbb{R}$ is the total degree count of atom $v$. By introducing such structural signals as strong prior knowledge, the attention score in **Supplementary Eq. (7)** can capture both semantic and space domain information.

*Multi-sense and multi-scale bond embedding.* Equipped with topological embedding, MechRetro is improved because beneficial structure messages are passed for better graph understanding. But it still neglects abundant bond information, which makes the model unable to distinguish some isomers that share the exact degree count of each atom.

Therefore, it is necessary to introduce bond information to improve the graph expressivity of the model. In the implementation, we encode bond properties (e.g., $\sigma$ orbit, $\pi$ orbit, conjugated bond, etc.) which are then represented as adjacency matrices $\{A_i \in \{0,1\}\}_{i=1}^{K_b}$, $A_i \in \mathbb{R}^{N \times N}$ of different $K_b$ senses, rather than using the type-based approach to embedding edge features. The reason is that by directly exposing specific reaction-related attributes (e.g., whether in the ring, etc.) hidden by bond type, MechRetro can better understand complex reaction data. Besides, some bond types share the same properties, e.g., C-C and C=C have $\sigma$ orbit, but conventional encoding for bond type cannot express such signal while multi-sense embedding is free of this message gap.

With bond encoding improved, the next problem is how to integrate it into graph data for a Transformer. It is not easy because an atom-pair-based bond cannot be embedded to a degree that a single atom defines. Inspired by **Eq. (2)**, where atom-pair-based semantic edges are learned, we consider bond information as natural prior knowledge of the molecular structure and add it to the semantics score before $SOFTMAX(\cdot)$. Then we have:

$$m_a^{(l)} = \sum_{u \in \mathcal{N}_v} \tilde{a}_{uv} h_u^{(l-1)} W_V , \qquad m_v^{(l)} = \tilde{a}_{vv} h_v^{(l-1)} W_V, \tag{5}$$

$$a_{uv} = \frac{h_u^{(l-1)} W_Q \left(h_v^{(l-1)} W_K\right)^T}{\sqrt{d_k}} + \sum_i^{K_b} \phi_i(A_{i;uv}), \qquad \tilde{a}_{uv} = SOFTMAX(a_{uv}), \tag{6}$$

where $\mathcal{N}_v = \mathcal{V}/\{v\}$ is a set of neighbors of atom $v$. However, it still cannot be as expressive as a simple GNN in practice, which excessively focuses on 1-hop neighbors in the space domain. To alleviate the problem, Graphormer[30] proposes a shortest-path embedding to acquire a global scope and achieves state-of-the-art (SOAT) performances in graph domain tasks. Meanwhile, a global distance can be easily calculated by bond length or 3d conformation optimized by MMFF[31]. Introducing global embedding, we have:

$$a_{uv} = \frac{h_u^{(l-1)} W_Q \left(h_v^{(l-1)} W_K\right)^T}{\sqrt{d_k}} + \sum_i^{K_b} \phi_i(A_{i;uv}) + RBF(A_{g;uv}), \tag{7}$$

where RBF is a gaussian radial basis function calculated as follows:

$$RBF(A) = \frac{exp(-(A - \phi_{mean})^2 / 2\phi_{std}^2)}{\sqrt{2\pi}\phi_{std}}. \tag{8}$$

It should be noted that global embedding introduces spatial information but ignores bond features in the overall scale. Besides, atom environment (AE), which regards several atoms as a token, plays a significant role in retrosynthetic prediction. Therefore, inspired by RetroTRAE[21], we propose a multi-level AE embedding that captures various radii of AEs by simply calculating different powers of the 0-1 adjacency matrix $A_i$. Thus, the final attention score is as follows:

$$bias_{uv} = \sum_i^{K_b} \sum_{j=1}^{K_r} \phi_i(A_{i;uv}^j) + RBF(A_{g;uv}), \tag{9.1}$$

$$a_{uv} = \frac{h_u^{(l-1)} W_Q \left(h_v^{(l-1)} W_K\right)^T}{\sqrt{d_k}} + bias_{uv}, \tag{9.2}$$

where $A_i^j$ is the j-th power of $A_i$, which describes AE with j radius, $K_r$ is the max radius of

AEs we take into account. Note that we later use the local item to denote $\phi_i(A_{i;uv}^j)$ and global item to denote $RBF(A_{g;uv})$. What calls for special attention is that $A_i^j$ with $A_i \in \{0,1\}^{N \times N}$ records the number of paths for length $j$ between each atom, which describes $j$ radius environment for each central atom. Finally, in **Eq. (9)**, we firstly employ a self-attention mechanism to learn a semantic relation for the combined feature of node, secondly introduce a multi-level environment $\{A_i^j\}_{j=1}^{K_r}$ and multi-sense embedding $\{A_i\}_{i=1}^{K_b}$ to take advantage of abundant bond information on a local scale, and thirdly, add a global embedding for atoms to capture a global scope in the spatial domain. Therefore, by calculating $a_{uv}$ in **Eq. (9)**, the expressive of MechRetro is at least as powerful as GNNs.

**Decision units and decision-making module**

In MechRetro, three decision units are designed for different demands, of which their outputs are all useful to retrosynthetic analysis for the planning of reaction pathways. In detail, the reaction center predictor (RCP) gives a probability distribution for each bond change which decides reaction centers, leaving group matcher (LGM) evaluates a score of compatibility between the product and each candidate leaving group, and leaving group connecter (LGC) determines the location where reaction center and leaving group connects. Notice that all decision units are closely related, so they are modeled in a joint distribution manner instead of in a separate training mode in previous work. Thus, the distribution of the three units can be modeled as:

$$f_\theta\left(\{\mathcal{G}_{r;i}\}_{i=1}^{N_r} \big| \mathcal{G}_{p;m}\right) = \underbrace{h_\theta\left(\{\mathcal{G}_{r;j}\}_{j=1}^{N_r} \big| \mathcal{G}_{p;m}, \{\mathcal{G}_{l;j}\}_{j=1}^{K_l}\right)}_{LGC} \underbrace{g_\theta\left(\{\mathcal{G}_{c;i}\}_{i=1}^{K_c}, \{\mathcal{G}_{l;i}\}_{i=1}^{K_l} \big| \mathcal{G}_{p;m}\right)}_{RCP\&LGM} \quad (10)$$

where $K_c, K_l$ denotes the number of reaction centers and leaving groups, respectively. Notice that $g_\theta(\cdot)$ in **Eq.(10)** also obscurely models the condition distribution between the given main product and unknown by-products. Because once the by-products are predicted correctly, the related reactants can be easily influenced by dual models of higher performance synthesis prediction.

*Reaction center prediction.* Instead of directly predicting the bond type of reaction centers, we factorize it into two sub-tasks for bond (RCP-B) or hydrogen (RCP-H) change. Between the two sub-tasks, we model the bond change prediction as a sparse edge link identification task and the hydrogen change prediction as node-level classification with $2k + 1$ labels, where $k$ is the max number of changed hydrogens. To predict reaction centers (RCs) for a given product $\mathcal{G}_p$ with node representation $h_v^{(L)}$ after $L$ layers, we calculate the probabilities of bond change $p_{uv}$ for atom pair $(u, v)$ and hydrogen change $p_v$

for atom $v$, which can be as follows:

$$p_{uv} = \sigma\left(CONCAT\left(\left\{\frac{h_u^{(L)}W_Q\left(h_v^{(L)}W_K\right)^T}{\sqrt{d_k}}\right\}_{i=1}^{n_{head}}\right)W_{bond}\right), \tag{11}$$

$$p_v = SOFTMAX\left(h_v^{(L)}W_{atom}\right), \tag{12}$$

where $W_{bond}, W_{atom}$ denote linear layers that aggregate messages from different heads. In addition, RCP is optimized as follows:

$$\mathcal{L}_B = -\sum_{(\mathcal{G}_p, RC)}\sum_{(u,v)\in RC} y_{uv}\, log(p_{uv}), \quad \mathcal{L}_H = -\sum_{(\mathcal{G}_p, RC)}\sum_{v\in RC} log(p_{v;y_v}). \tag{13}$$

*Leaving group matching.* We model leaving group matching as a graph-level multi-classification task instead of an autoregressive manner. The reason is that we find a small ratio (about 0.46%) between the leaving group and the number of reactions, meaning several definite patterns appear in leaving groups. As a result, after applying a heuristic breadth-first traversal algorithm to unify permutations for nodes in the same kind of leaving groups, we can obtain 228 types of leaving group in USPTO-50K. Based on these collected leaving groups, we construct a vocabulary $\mathcal{V}_{LG}$ whose index maps a determinate leaving group. Therefore, the predicted distribution $p_{\mathcal{G}_p}$ and optimizing target for LGM are as follows:

$$p_{\mathcal{G}_p} = h_{\mathcal{G}_p}W_{LG}, \quad h_{\mathcal{G}_p} = h_s^{(L)}, \quad \mathcal{L}'_{LG} = -\sum_{\mathcal{G}_p} log\frac{exp(p_{\mathcal{G}_p; y_+})}{\sum_{y\in\mathcal{V}_{LG}} exp\left(p_{\mathcal{G}_p; y}\right)}, \tag{14}$$

where $h_{\mathcal{G}_p}$ is a graph-level representation for $\mathcal{G}_p$, $h_s$ denotes feature of the super node. The super node is like $<cls>$ token in natural language processing and is a virtual node in a practical sense, which is set to be connected by a virtual edge to all the other nodes.

*Leaving group connection.* We also design LGC as a sparse edge link process like RCP-B in **Eq. (11)**. The difference is that we only predict the connections between connected atoms in LGs and product, which reduces computation complexity from $O(NM)$ to $O(N)$, where $N, M$ denotes the number of atoms for product and LG, respectively. The motivation is the limited number of connected atoms in LGs.

*Mechanism-like decision process.* Different from end-to-end prediction like previous work, MechRetro can provide a more transparent and interpretable prediction for various retrosynthesis analysis demands. As shown in **Fig.1f**, the whole process can be divided into five actions: leaving group matching, initializing, leaving group connecting, bond changing, and hydrogen changing. The process is evaluated by a flexible user-designed energy function of the predicted probability distribution. In leaving group matching, the energy score is associated with compatibility between leaving group and the product through LGM. In initializing, the selected leaving group is not attached, and all the potential reaction centers keep intact. In leaving group connecting, the connected bonds are identified and scored by LGC. Notably, the information on connected bond type is recorded in a matched leaving group. In bond changing, the bonds from reaction centers are identified and scored by RCP. At the hydrogen changing, we adjust the amount of hydrogen for the noncompliant atoms and filter out all the illegal molecules to generate final reactants. In this paper, the energy function is defined as the negative log function of the probabilities of associated conditions.

**Self-adaptive joint learning module**

***Contrastive learning enhanced leaving group matching.*** In LGM, given a product $\mathcal{G}_p$, we directly predict the index of the most appropriate leaving group for $\mathcal{G}_p$, which, however, still ignores the rich structure information of leaving group and limits the performance of overall prediction. Hence, benefiting from maturing contrastive learning techniques in molecular representations (e.g., MolCLR[32], 3D-infomax[33], etc.), we adopt a strategy that minimizes the similarity of representations between leaving group $h_{\mathcal{G}_{LG}}$ and matched product $h_{\mathcal{G}_p}^+$ and maximize it in any other conditions. Based on this supervised contrastive-learning-based idea, an optimizing object inspired by SupContrast[34] be can as follows:

$$\mathcal{L}'_c = -\sum_{\mathcal{G}_{lg}} \frac{1}{|\mathcal{G}_p^+|} \sum_{\mathcal{G}_p^+} \log \frac{exp\left(sim\left(h_{\mathcal{G}_{lg}}, h_{\mathcal{G}_p}^+\right)/\tau\right)}{\sum_{\mathcal{G}_p} exp\left(sim\left(h_{\mathcal{G}_{lg}}, h_{\mathcal{G}_p}\right)/\tau\right)}, \quad (15)$$

where $sim(h_u, h_v)$ measures the similarity between $h_u$ and $h_v$, $\tau$ is a hyper-parameter and is a scalar. In practice, recognizing a positive instance for every batch brings in extra costs. So, after regular forward of LGM, we decorate every $h_{\mathcal{G}_{LG}}$ with the same contrastive token and then input it to the LGM to predict the corresponding index in vocabulary $\mathcal{V}_{LG}$. Thus, the enhanced loss function $\mathcal{L}_{LG}$ for LGM is as follows:

$$\mathcal{L}_{lg} = -\frac{1}{2}\left(\sum_{\mathcal{G}_p} \log \frac{exp\left(p_{\mathcal{G}_{p;y_+}}/\tau\right)}{\sum_{y \in \mathcal{V}_{lg}} exp\left(p_{\mathcal{G}_{p;y}}/\tau\right)} + \sum_{\mathcal{G}_{lg}} \log \frac{exp\left(p_{\mathcal{G}_{lg;y_+}}/\tau\right)}{\sum_{y \in \mathcal{V}_{lg}} exp\left(p_{\mathcal{G}_{lg;y}}/\tau\right)}\right), \quad (16)$$

where $p_{\mathcal{G}_{LG}} = (h_{\mathcal{G}_{LG}} + h_{con})W_{LG}$ denotes the distribution that LGM predicts, $h_{con}$ is the feature of the contrastive token. This strategy plays the same role as **Eq. (14)** without the above extra costs. Besides, what needs to be emphasized is that we only apply this technique in the training stage. As a result, we can make MechRetro perceive structural information of leaving groups without extra prior knowledge.

***Self-adaptive joint learning.*** To predict probabilities of bond changes $p_{uv}$, hydrogen attachments $p_v$, and leaving group matching $p_{LG}$ jointly, product $\mathcal{G}_p$ is first sent into shared layers to learn the mutual representation of three tasks, which then is passed to each specific task layer. Because of the immediate relevance between the reaction center and leaving group, it can be a mutual promotion learning process for joint training. However, optimizing three targets at the same time can be conflicting for parameters in a shared layer due to the discordance of complexity for tasks and magnitude for loss functions. In MechRetro, we propose a multi-task learning strategy that can adaptively adjust weights for the above three losses. In detail, we introduce a decent rate for $i$-th loss $r_i^{(t)}$ in $t$-th training step to measure the complexity of $i$-th task and a normalizing coefficient $\alpha_i$ to unify the magnitude of $i$-th loss. Combined above, the total loss of $t$-th step $\mathcal{L}_T^{(t)}$ is as follows:

$$\mathcal{L}_T^{(t)} = \sum_{i=1}^{K_t} \frac{exp\left(r_i^{(t)}/\tau\right)}{\sum_{j=1}^{K_t} exp\left(r_j^{(t)}/\tau\right)} \alpha_i \mathcal{L}_i^{(t)}, \quad r_i^{(t)} = \frac{\mathcal{L}_i^{(t-1)}}{\mathcal{L}_i^{(t-2)}}, \quad \alpha_i = \frac{1}{\mathcal{L}_i^{(0)}}. \quad (17)$$

## Results and discussions

**Our MechRetro outperforms the state-of-the-art approaches**
To evaluate the effectiveness of MechRetro, we compared it with several state-of-the-art retrosynthesis approaches on USPTO-50K[35], a commonly used benchmark dataset in retrosynthetic work. It contains 50,000 reactions marked with ten reaction types. For a fair comparison, we adopted the same split scheme following the previous work[26,29,36].

Table 1 presents the prediction results of our MechRetro and other existing approaches in terms of the top-$k$ exact match accuracy with $k \in \{1,3,5,10\}$. Observations are as follows. (1) Compared to other retrosynthetic Transformer-based models (decorated with superscript T), our MechRetro architecture achieves the best performance in conditions both reaction types known and unknown. Specifically, MechRetro remarkably outperforms the runner-up Transformer-based method - G2GT[43], respectively achieving the improvement by 3.9%, 3.6%, 2.7%, and 1.7% in terms of top-1 accuracy, top-3 accuracy, top-5 accuracy, and top-10 accuracy with reaction unknown conditions. This demonstrates that our MechRetro is more effective than existing Transformer-based methods in the retrosynthesis prediction task. (2) In comparison with those template-based models (marked with a double asterisk in Table 1) and semi-template-based models (marked with a single asterisk in Table 1), we can observe consistent results; that is, our MechRetro is superior to the approaches. The possible reason is that our template-free MechRetro sufficiently captures the structural information of leaving group as prior knowledge, which helps to significantly improve the predictive performance. This demonstrates that integrating leaving group to build the predictive model can introduce more effective information than those approaches leveraging reaction templates for model training.

Table 1. Performance of our MechRetro and the state-of-the-art methods on USPTO-50K benchmarks.

| Model/Task | Top-k Accuracy (%) | | | | | | | |
| --- | --- | --- | --- | --- | --- | --- | --- | --- |
| | Reaction class known | | | | Reaction class unknown | | | |
| k = | 1 | 3 | 5 | 10 | 1 | 3 | 5 | 10 |
| **Overall Prediction** | | | | | | | | |
| **[T]RetroSim[36] | 52.9 | 73.8 | 81.2 | 88.1 | 37.3 | 54.7 | 63.3 | 74.1 |
| **[G]NeuralSym[37] | 55.3 | 76.0 | 81.4 | 85.1 | 44.4 | 65.3 | 72.4 | 78.9 |
| **[G]GLN[29] | 64.2 | 79.1 | 85.2 | **90.0** | 52.5 | 69.0 | 75.6 | **83.7** |
| *[G]G2Gs[24] | 61.0 | 81.3 | 86.0 | 88.7 | 48.9 | 67.6 | 72.5 | 75.5 |
| *[G+T]RetroXpert[26] | 62.1 | 75.8 | 78.5 | 80.9 | 50.4 | 61.1 | 62.3 | 63.4 |
| *[G]GraphRetro[13] | 63.9 | 81.5 | 85.2 | 88.1 | 53.7 | 68.3 | 72.2 | 75.5 |
| [T]SCROP[38] | 59.0 | 74.8 | 78.1 | 81.1 | 43.7 | 60.0 | 65.2 | 68.7 |
| [T]LV-Transformer[19] | - | - | - | - | 40.5 | 65.1 | 72.8 | 79.4 |
| DualTF[39] | 65.7 | 81.9 | 84.7 | 85.9 | 53.6 | 70.7 | 74.6 | 77.0 |
| [T]GET[40] | 57.4 | 71.3 | 74.8 | 77.4 | 44.9 | 58.8 | 62.4 | 65.9 |
| [T]AutoSynRoute[41] | - | - | - | - | 43.1 | 64.6 | 71.8 | 78.7 |
| [T]TiedTransformer[42] | - | - | - | - | 47.1 | 67.1 | 73.1 | 76.3 |
| [GT]G2GT[43] | - | - | - | - | 54.1 | 69.9 | 74.5 | 77.7 |
| [T]Graph2SMILES[44] | - | - | - | - | 52.9 | 66.5 | 70.0 | 72.9 |
| [T]RetroPrime[45] | 64.8 | 81.6 | 85.0 | 86.9 | 51.4 | 70.8 | 74.0 | 76.1 |
| [G]GTA[46] | - | - | - | - | 51.1 | 67.6 | 73.8 | 80.1 |
| $^{GT}$MechRetro (**ours**) | | 68.2 | 83.7 | 87.2 | 89.0 | 58.0 | 73.5 | 77.2 | 79.4 |

Top-k accuracy on overall, reaction center prediction, and synthon completion tasks when reaction type both known and unknown compared to other retrosynthesis models. Among them, template models are marked with a double asterisk, while semi-template models are marked with an asterisk. Besides, Transformer-based models are marked with T, while graph-based models are marked with G.

**Uncertainty estimation for retrosynthetic reasoning process and query evaluations**

MechRetro can provide multidimensional energy scores for uncertainty estimations of the predictions and queries. As shown in **Fig. 2**, apart from the overall energy as a soft evaluation that assesses the rationality of the predicted reactants, four energy differences can also be obtained to measure the uncertainty of identified leaving group, connection conditions, bond changes, and chemical fitness of structure. Notice that, for the credible predicted reactants, the energy value goes through a process of first increasing and then decreasing. Therefore, the process is more like a reversed reaction mechanism process. Interestingly, the state in initializing is similar to the transition state. For the query prediction in Fig. 2, the input product needs to cross a higher energy barrier, which means the leaving group of the query prediction doesn't match the input product. Meanwhile, the energy value doesn't go through a decent stage, illustrating that the connection and reaction center is inappropriate for the product. And finally, the energy evaluation of the query product is up to 86.72, i.e., an unreasonable prediction.

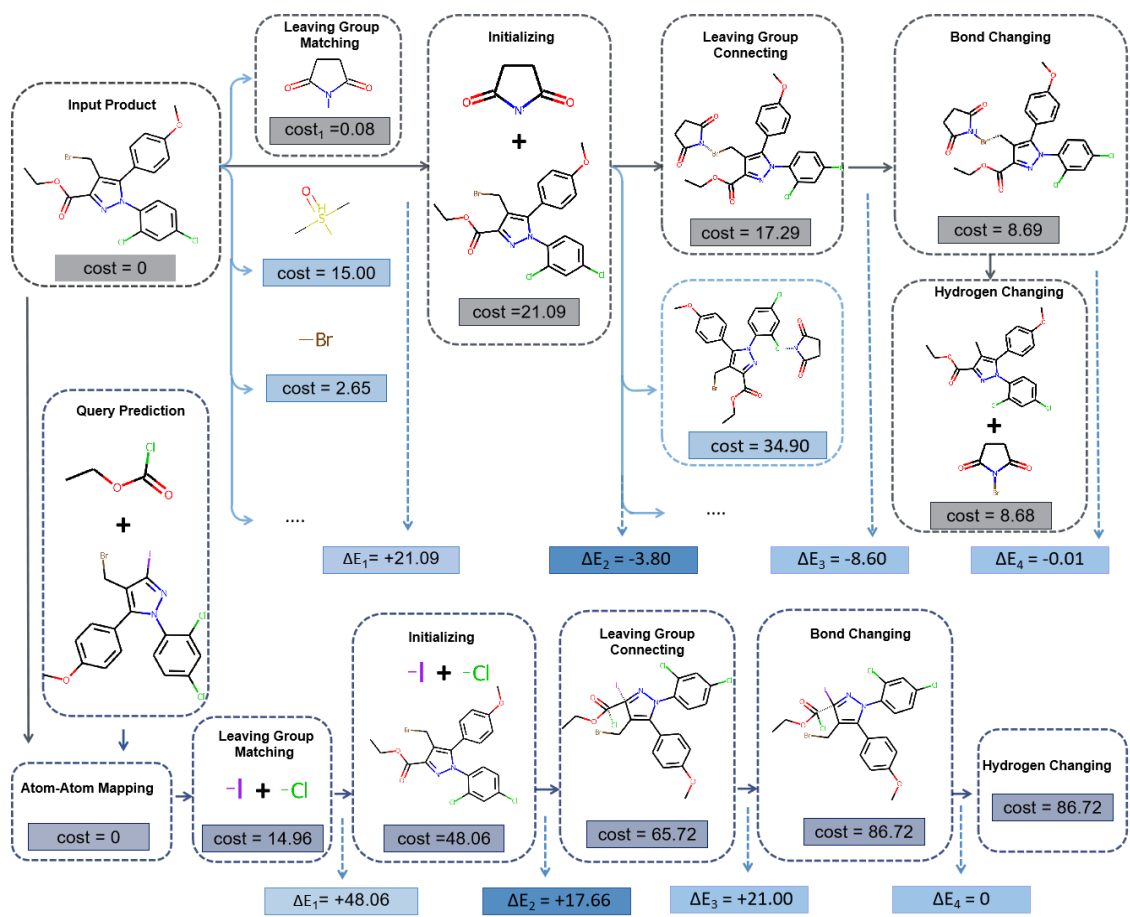

**Fig.2. Uncertainty estimation for reasoning process and query evaluations.** Four energy scores $\Delta E_i, i \in \{1,2,3,4\}$ are used to describe the reliability of LGM, RCP, LGC, etc.

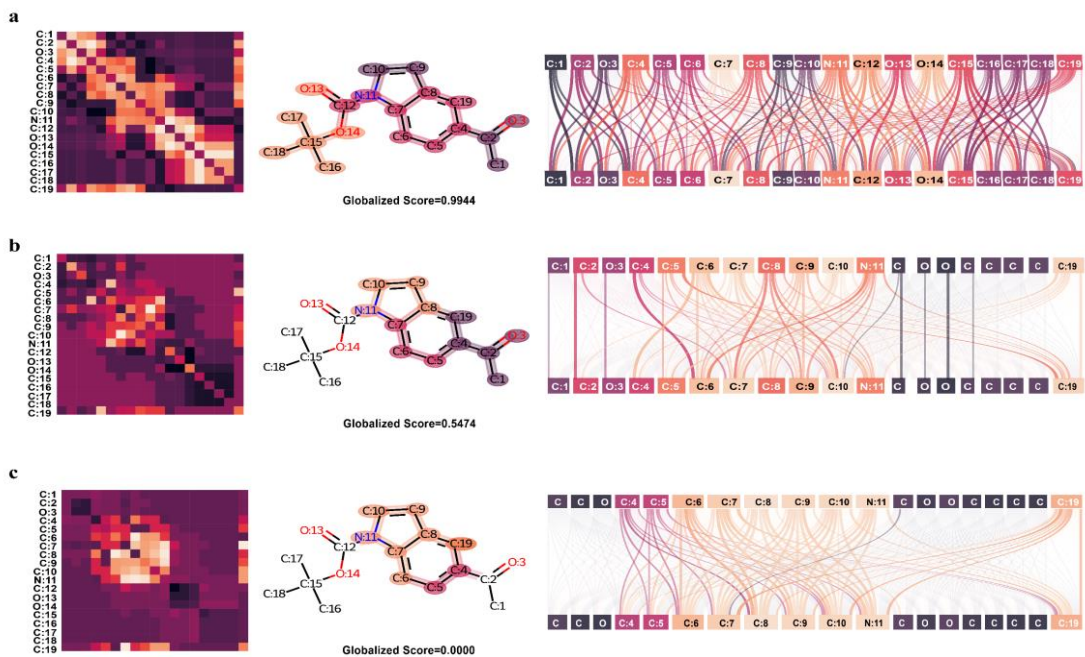

**Fig. 3. Heat map of attention bias.** The j-hop matrices are normalized. The j-th power of the adjacent matrix can be used to measure the topological relation for the j-hop atom pairs in a molecule. The RV coefficient sorts the heatmaps of attention bias with the global item matrix in descending order. a. global-item-dominated heatmap. b. local-item-dominated heatmap. c. mixed-item-dominated heatmap.

**Multi-scale expressivity on edge representation**

To figure out the expressivity on the edge feature of MechRetro, we extract attention bias in **Eq. (9.1)** as adjacent matrices representing edge features for the sampled molecular graph from different self-attention heads. We showcase three kinds of hot maps as shown in **Fig.3** and for more in **Supplementary Fig.1**. Compared to the adjacent matrix of naive bond type (see **Supplementary Fig.1a**), the proposed bond embedding captures a richer scale of topological edge information (see **Supplementary Fig.1b**), in which each head assigns a different level of attention bias to atoms in the molecule graph. Notice that, in these adjacent matrices from various heads, some of the patterns appear as a diagonal tree, which means these globally focused heads are dominated by global item, while other patterns consist of squares with diverse sides length, illustrating the locally focused heads are controlled mainly by local item. To further demonstrate this point, we calculate the RV coefficients[47] between attention biases and global items to measure the two types of matrix similarity and then sort the heatmaps of attention bias in descending order. As shown in **Fig.3**, according to their value of RV coefficient, the heatmaps of attention bias can be divided into three classes: 1) global-item-dominated heatmap with RV coefficient more prominent than 0.90 and the pattern tending to the diagonal tree; 2) local-item-dominated heatmap with a zero RV coefficient and a squares-like pattern; 3) mixed-item-dominated heatmap whose RV coefficient ranges from 0.61 to 0.50 and pattern appears both diagonal tree and squares. Therefore, the heatmaps illustrate that the multi-head bond embedding block is capable of noticing multi-scale molecular structure information and, in the meantime, is more explicable than the edge feature concatenation method applied by conventional GNNs on graph edge representation.

**Ablation studies on scale information and augmentation strategy**

To further explore the influence of attention bias on the results, we conduct an ablation study based on scale information (see **Fig.4a-c**). As shown in **Fig.4a** and **Fig.4b**, the top1 accuracy increases and decreases as the max hop grows up (from 0 to 5). Besides, the accuracy reaches its maximum when the max hop reaches 4. This is most likely because of the competition between global and local items. When max hop decreases to a relatively small positive integer, which means the local item in **Eq.(9.1)** missing more high order items than before, the attention biases of different heads tend to neglect more local scale structure, resulting in the decreasing of top1 accuracy, and vice versa. Thus, when max hop equals four, the local and global items reach a trade-off and their sum, attention bias, takes multi-scale structural characteristics into account. In addition, the global item appears to be more important than the local item. Compared to the maximal point (max hop equals four), the decline is more pronounced when the global item is masked than that when the local item is set to zero. Furthermore, when the global item is masked, the local item has a rising influence on the outcome performance, which illustrates that there exists an overlap in the embedded structural information that the two types of items provide.

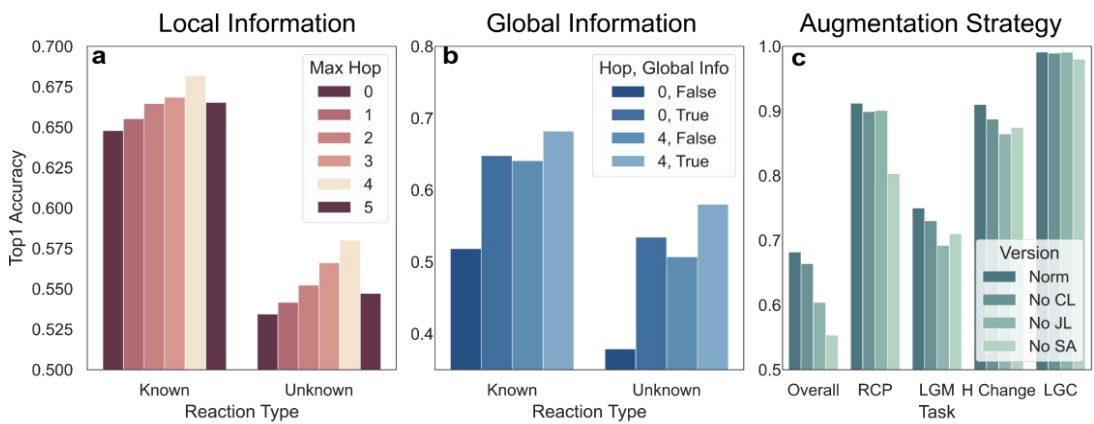

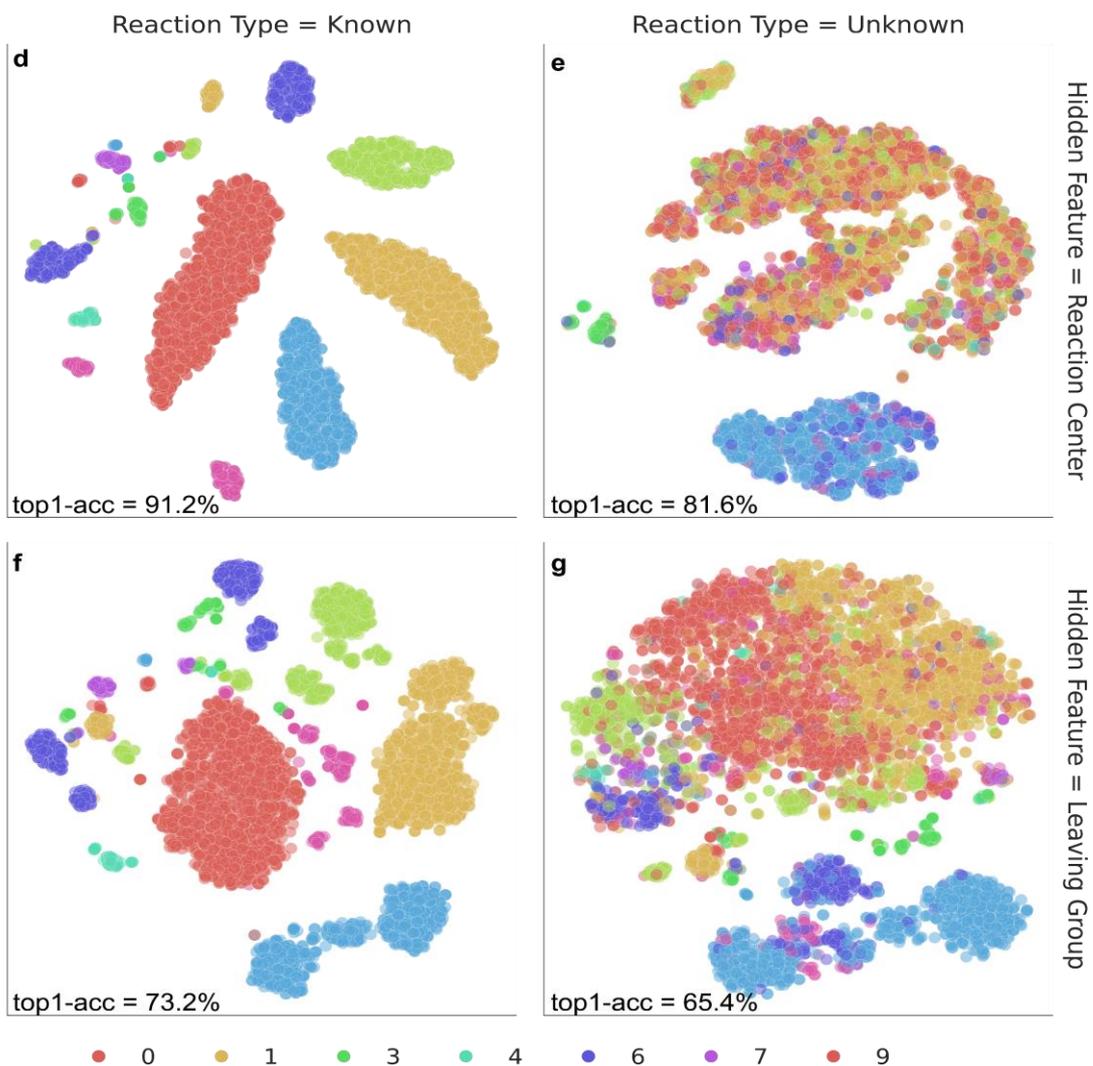

**Fig.4. Ablation studies and distributions of t-SNE from MechRetro hidden layer. a and b. Ablation study on scale information with reaction type known and unknown conditions.** We mainly focus on max hop of local item. With max hop changing to 0 or 4, we remove the global item to verify the validity. With the value of max hop arising, the top1 accuracy increases and decreases. When max hop equals 4, the top1 accuracy reaches maximum. c. **Ablation study on different augment strategies for overall prediction and three subtasks.** We consider four conditions to compare: 1) normal version and max hop set to 4, 2) removing contrastive strategy, 3) not adopting

adaptive coefficient in multi-task learning, 4) removing all shared layers and training each subtask individually. **d. Distributions of t-SNE from hidden layers of MechRetro.** Hidden features are extracted according to 1) whether the reaction type is known, 2) and where are the hidden features from (RCP layer or LGM layer), which are then compressed into two dimensions by t-SNE tools. Different colors denote different reaction types.

To figure out the contributions of three augmentation strategies. We also conducted an ablation study on the two augment strategies, which can be divided into four versions: norm, no contrastive learning (CL), no joint learning (JL), and no self-adaptive learning (SA), shown in **Fig.4c**. By comparing the normal and no CL versions, we can demonstrate that the contrastive training process in MechRetro may help the model perceive the structure information of leaving group and improve the LGM task by about 5% top-1 accuracy successfully. And even the RCP and LGC are also strengthened. It is mainly due to the propagation of shared knowledge from the LGM layer to other layers. As for self-adaptive multi-task learning, we remove all the self-adaptive coefficients in **Eq. (16)** and sum all the sub-losses. From **Fig.4c**, the no SA version is lower than any other version in most tasks. It is mainly a result of the improper coefficients, which cannot balance the sub-tasks with varying difficulties well. Notice that, as an alternative to the standard version, a different group of hyperparameters can be introduced for each task as fixed coefficients of the corresponding loss functions. However, this alternative raises two significant problems: 1) the optimum of the combination of coefficients is hard to search, and 2) even if the optimum of coefficients is found, it may not fit every epoch in the training stage dynamically as fixed hyperparameters. Our adaptive coefficients in MechRetro are necessary for the multi-task framework to avoid the two problems. In addition, we also remove the whole joint learning framework in MechRetro directly and train the sub-tasks individually. The no JL version outperforms others in some easier tasks (e.g., LGC task) due to the less constraint that the LGC layer imposes on learnable parameters. However, because of the absence of joint learning, the shared knowledge learned from easier tasks cannot be transmitted to other harder tasks, which in turn triggers the degradation of performance for the harder tasks. So, we can demonstrate both contrastive training and self-adaptive joint learning can improve the performance of MechRetro.

**Influence of reaction type**

Different from reaction type unknown circumstance, when reaction type is given, we add extra embeddings into a super node which is then extracted as graph-level representation after $L$ message-aggregation layers. **Table 1** shows the increase of top-k accuracy for introducing the message of reaction type. To investigate how reaction type token affects the performance of MechRetro, we extract four kinds of hidden features according to their source (from last RCP layer or last LGM layer) and the condition of being informed with reaction type. The labels of reaction type color the distributions of the compressed hidden features through t-SNE in **Fig.4b**. Obviously, reaction type message imposes a more regular constraint for hidden representation of product than that of being free from reaction type limits, which thus enhances the location process of reaction centers and matching procedure of leaving groups. Besides, on the one hand, we also find different clusters in the same reaction type from **Fig. 4d** and **Fig. 4f**, which means these reaction types can further be divided into more advanced classes given the specific task.

On the other hand, when the reaction type is not given (**Fig. 4e**, **Fig. 4g**), we can also obtain a rough outline relating to it, which generalizes the similarities between different classes on the task perspectives. Notably, what is interesting is that the more discrete the distribution is, the lower the accuracy of the task (from 91.2% to 81.6% on RCP and from 73.2% to 65.4%). Therefore, to improve the performance of the reaction type unknown model, an effective way is to impose extra constraints for hidden features only during when training stage (e.g., SupContrast[48] and our contrastive technic for leaving groups). These constraints improve the fitting ability within the domain of the training sets. However, the extra regular loss functions also increase the risk of overfitting, which means the degradation of generalization to samples out of the domain.

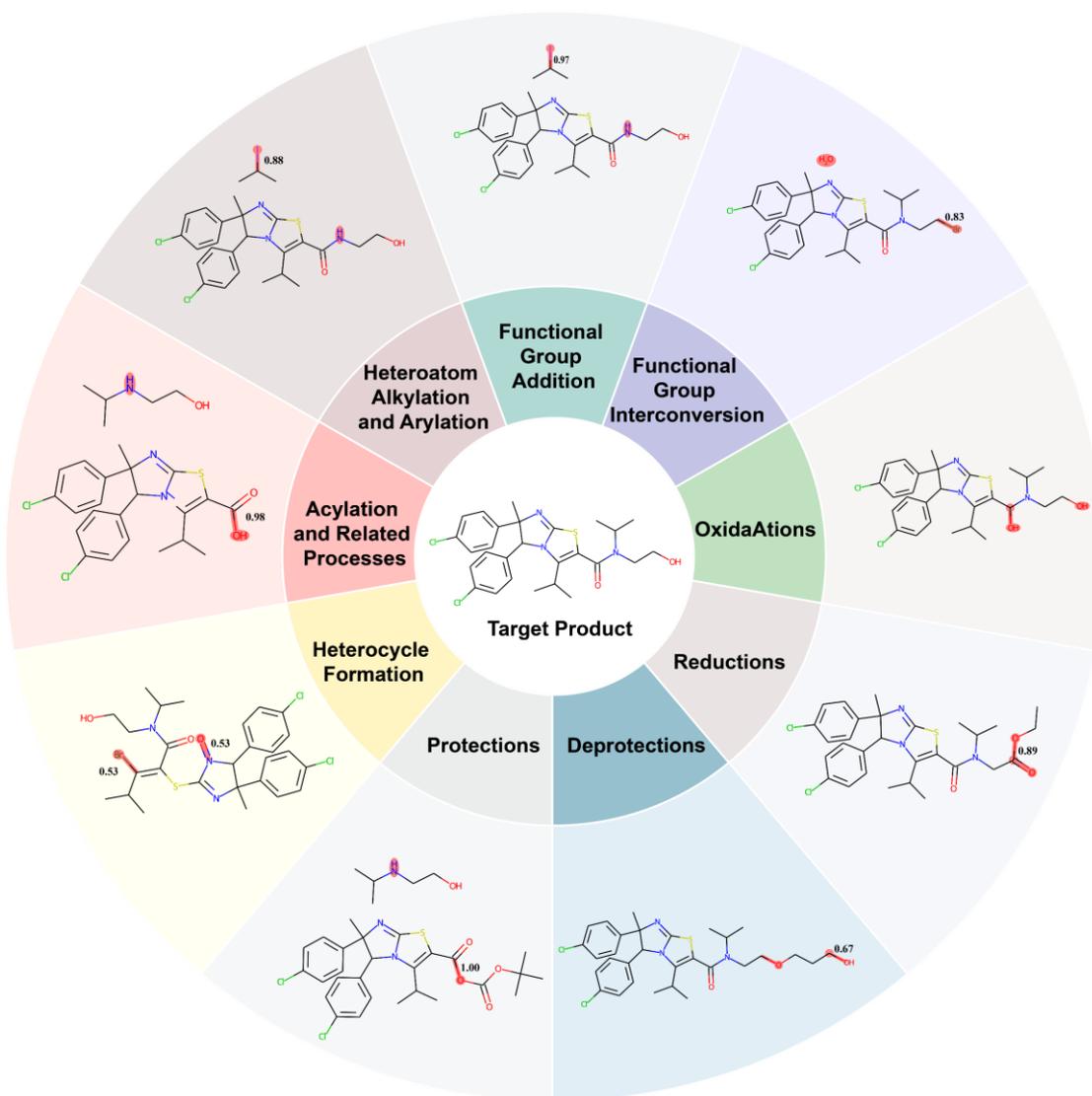

**Fig.5. Reaction type mutation experiment.** Randomly sampled product from the test set is added by different indices of reaction type embeddings and then resent into MechRetro. Although most cases do not conform to the distribution of the initial cases, MechRetro still performs well, and the predictions are consistent with the corresponding reaction type.

**Reaction type mutation experiment and directed retrosynthesis prediction**

As mentioned above, reaction type embeddings enhance the prediction performance by imposing constraints on features of hidden layers. However, this strong prior knowledge also can be a significant obstacle to making a feasible prediction when the target reaction type deviates from the distribution of the training set. To alleviate the limitation and demonstrate the generalization of MechRetro, we randomly sample several cases from the test set and screen out an appropriate case to meet the requirements of various reaction types (e.g., heterocycle for reaction type of heterocycle formation). Then, we mutate the reaction type of the screened sample into other different types and reinput the example into MechRetro for predictions. **Fig.5** shows that the predicted reactants which are out of the distribution of the dataset domain can be consistent with the corresponding reaction type, which means our MechRetro grasps a good understanding of reaction types. Meanwhile, from an application view, with the directed reaction type and product molecule being input, MechRetro can predict robust and directed reactants to meet various needs of users. In addition, our learned embeddings of reaction type are potential in other related tasks (e.g., encodings of reaction, organic mechanism inference, etc.)

**Atom-perturbation-based explainability (APEx) and reaction type tracing analysis.** To investigate the effect of substructure in a molecular on the results and how MechRetro understands each task, we adopt an APEx explainability algorithm shown in **Supplementary Algorithm 1**. To generalize, we record the initial loss of the given sample, mask the features of each atom and corresponding edges, calculate the change rates of the loss values, and finally get the contribution graph by the change rates. Notice that the score (for atom or edge) in the contribution graph directly describes how the loss will increase relative to the initial value if the corresponding substructure is masked. To make the contribution graph easier to understand, we extract several tasks for visualization, as shown in **Fig.6a**. **Fig.6a** informs the ester amide is the key functional group for sampled reaction to the RCP task while the easter group is the most significant substructure for LGM task. As for an overall prediction, we comprehensively consider the scale and variation span of the loss values of four subtasks. Finally, the ester amide is the key functional group to the sampled reaction.

As an extension, we combine APEx with the reaction type mutation experiment. In detail, given a product molecule with a reaction type label, APEx can return a contribution with input reaction type. Based on this process, we can fix the product while simultaneously mutating the label of reaction type in turn and input it separately into the APEx to get a contribution vector whose elements describe sensitivity to the focused atom for each mutated reaction type respectively. As shown in **Fig.6b**, we provide both a hard label version that only considers the mutated reaction type that the substructure focuses on most and a soft label version that mixes all mutated reaction types by the contribution vector. Both versions can indicate which type of reaction the focused substructure may come from for a particular task. E.g., in the RCP task with a hard label version, the hydroxyl is associated with reaction type five (deprotections), which means the RCP thinks the hydroxyl is most likely synthesized by deprotections reaction type. While in the LGM task, the hydroxyl is most likely synthesized by reaction type 8 (functional group interconversion). Since the suboptimum solution of other reaction types should not be neglected when the contribution score of different reaction types is close to the optimum, the soft label version is also necessary. In the soft label versions, the color of the substructure is calculated by weighted sum according to ten anchor colors and the contribution scores. From the overall prediction with the soft label version in **Fig.6b**, MechRetro focus on heterocyclic sulfur and highlights amine structure to synthesize the sampled product.

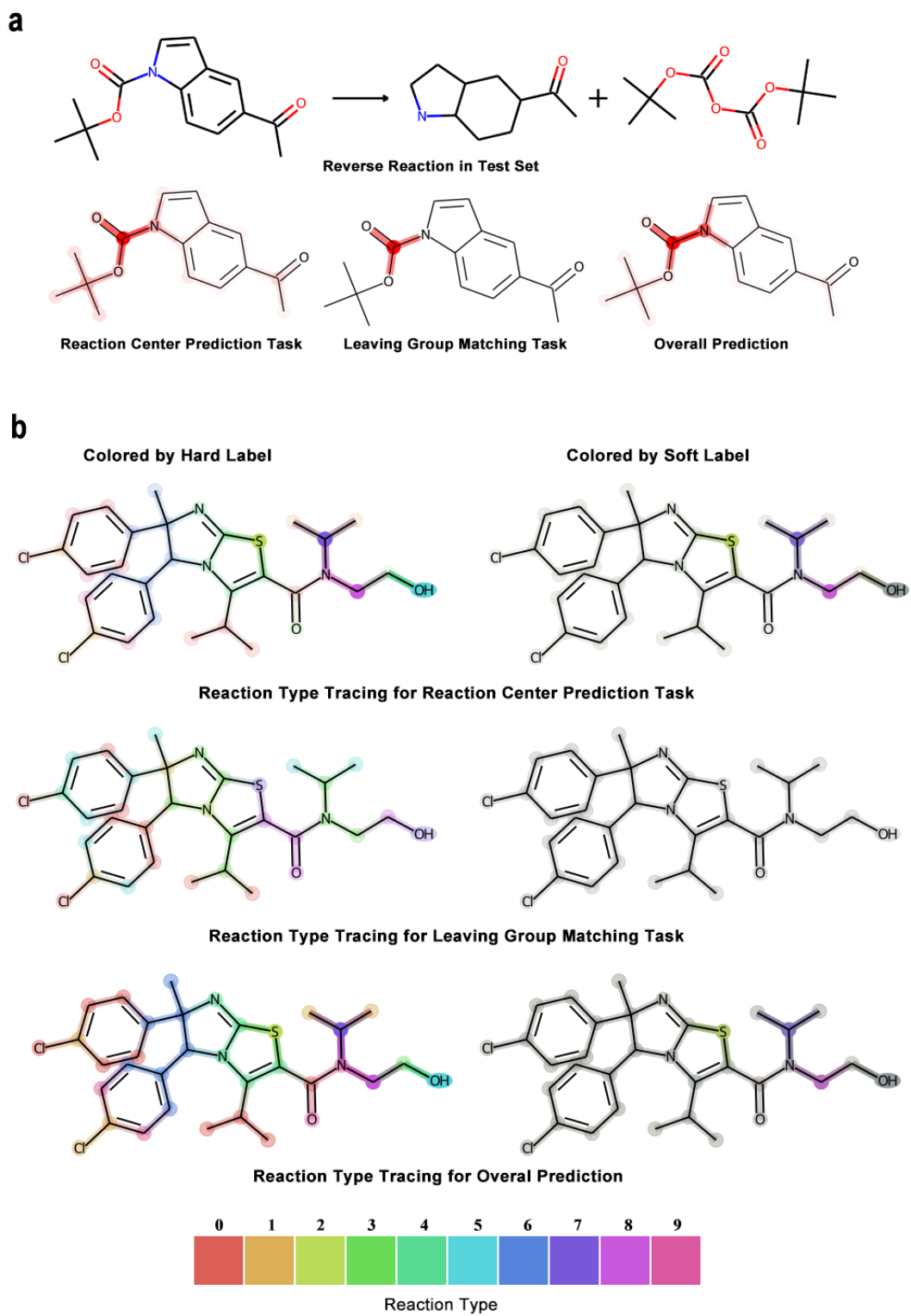

**Fig.6. Reaction type mutations. a. Sampled reverse reaction in the test set and extracted contribution graph obtained by atom-perturbation-based explainability algorithm.** From the contribution graph, MechRetro considers ester amide, ester, and ester amide as the critical substructure to RCP, LGM, and Overall prediction task, respectively, for the sampled reaction in the test set. **b. Reaction type tracing result for extracted three tasks.** For hard labels, we dye the molecule with the anchor color, representing the top-1 reaction type in the contribution vector for each atom. In contrast, for the soft labels, we use a mixed color whose RGB is weighted and summed by the contribution vector. The depth of the substructure's color indicates the structure's contribution to the corresponding inverse reaction.

## Extending to retrosynthesis pathway planning

Considering a more practical scenario of pathway planning, we integrate the Retro*[49] algorithm into our MechRetro, where our designed energy score replaces the cost evaluation. To show the effectiveness of hybrid MechRetro, we take protokylol (a β-adrenergic receptor agonist used as a bronchodilator) as an example. And as shown in Fig. 7, our MechRetro devises a four-step-synthetic-route for protokylol, in which the energy scores of the decision process illustrate the key sub-process to support the MechRetro makes the corresponding prediction. To further demonstrate the practicability of the proposed scheme, we conducted a literature search as evidence for each reaction step. Notice that the most proposed reactions cannot be searched, but we find similar reactions with high yield matching proposed reactions.

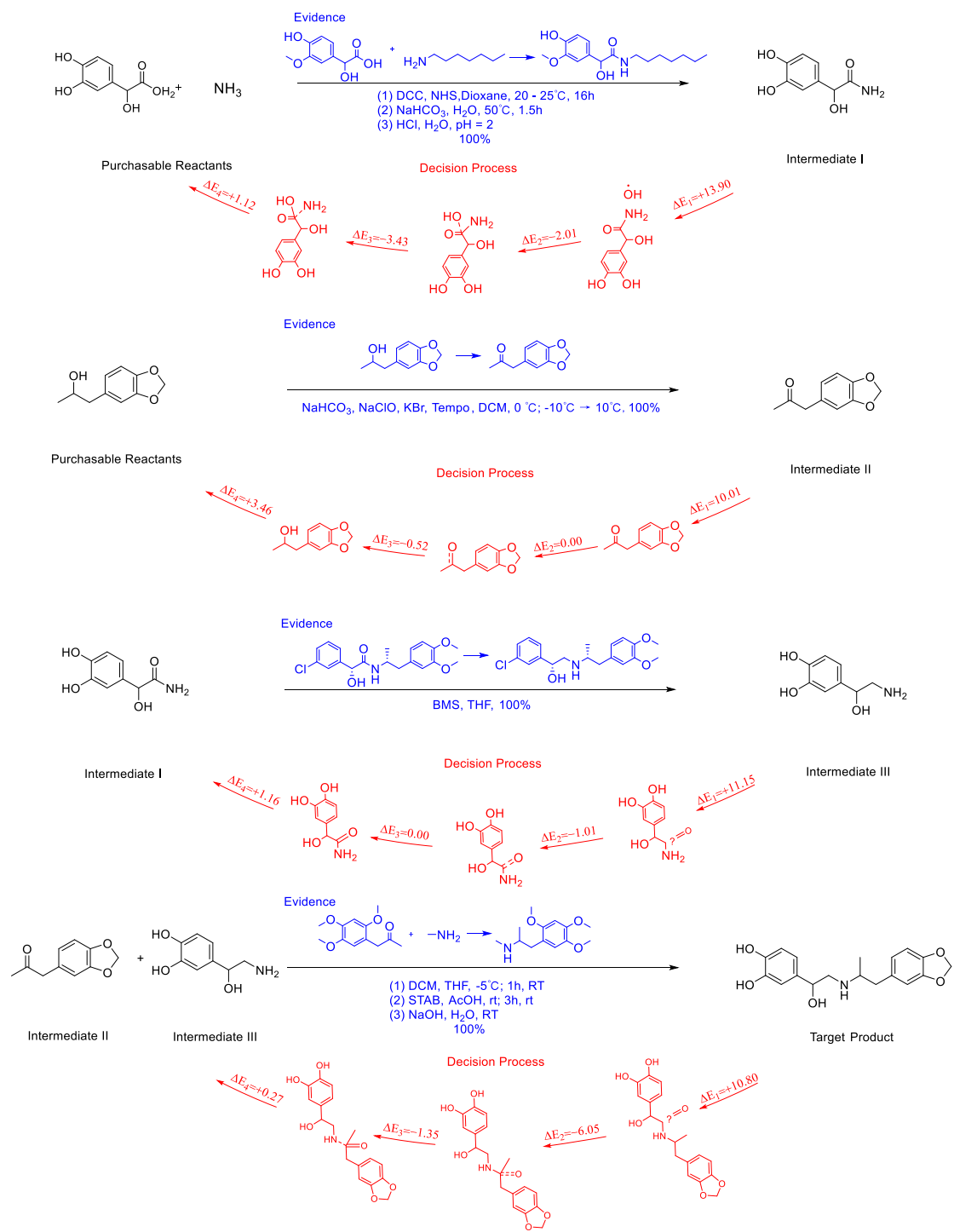

**Fig.7. Retrosynthesis pathway planning on protokylol.** MechRetro devises a four-step synthetic route for protokylol.

## Conclusion

In this study, we have proposed MechRetro, an interpretable pipeline that simulates the chemical mechanism process for retrosynthesis prediction and pathway planning. Our MechRetro uses a novel self-adaptive joint learning Graph Transformer that is enhanced by an efficient contrastive training strategy. The reasoning process of our MechRetro can be explained as a reaction path process with energy for uncertainty assessment.

Experimental results showed that (1) our MechRetro achieves the best performance for the single-step retrosynthesis prediction compared with the current popular sequence-based and graph-based models; (2) the proposed Graph Transformer combining multi-scale molecular structure information is proved to be a robust and powerful deep neural network for molecule representations; (3) the self-adaptive joint learning and contrastive learning framework could take full advantage of shared knowledge among related tasks, demonstrated to be potential tools for multi-objective optimization scenarios.

Furthermore, we employ the atom-perturbation-based explainability algorithm to reveal the important substructure for different sub-tasks and then combine the explainable algorithm with the reaction type mutation experiment to explore the critical substructure and corresponding reaction type. In addition, we also extend MechRetro to the interpretable multi-step scenario for further identifying efficient synthetic routes. Overall, we expect MechRetro to provide meaningful insights for high-throughput automated organic synthesis in drug discovery.

## Availability of data and materials

All code used in data analysis and preparation of the manuscript, alongside a description of necessary steps for reproducing results, can be found in a GitHub repository accompanying this manuscript: https://github.com/Wanyinee/MechRetro.

## Competing interests

The authors declare that they have no competing interests.

## Author contributions

Y.W., L.W. conceived the basic idea and designed the research study. Y.W. developed the method. Y.W., C. P., YZ. W, Y.J, S. L. JJ. R. drew the figure. Y.W. and L.W. wrote the manuscript.

## Funding

The work was supported by the Natural Science Foundation of China (Nos. 62072329, and 62071278).

## References

1    Corey, E. J. & Wipke, W. T. Computer-Assisted Design of Complex Organic Syntheses. *Science* **166**, 178-192, doi:doi:10.1126/science.166.3902.178 (1969).
2    Szymkuć, S. *et al.* Computer-Assisted Synthetic Planning: The End of the Beginning. *Angewandte Chemie International Edition* **55**, 5904-5937, doi:https://doi.org/10.1002/anie.201506101 (2016).
3    Mikulak-Klucznik, B. *et al.* Computational planning of the synthesis of complex natural

products. *Nature* **588**, 83-88, doi:10.1038/s41586-020-2855-y (2020).

4   Corey, E. J. The Logic of Chemical Synthesis: Multi-step Synthesis of Complex Carbogenic Molecules(Nobel Lecture). *Angewandte Chemie International Edition in English* **30**, 455-465, doi:10.1002/anie.199104553 (1991).

5   Corey, E. J. Robert Robinson Lecture. Retrosynthetic thinking—essentials and examples. *Chemical Society Reviews* **17**, 111-133, doi:10.1039/CS9881700111 (1988).

6   Corey, E. J. & Cheng, X. M. *The Logic of Chemical Synthesis*. (Wiley, 1989).

7   Coley, C. W., Barzilay, R., Jaakkola, T. S., Green, W. H. & Jensen, K. F. Prediction of Organic Reaction Outcomes Using Machine Learning. *ACS Central Science* **3**, 434-443, doi:10.1021/acscentsci.7b00064 (2017).

8   Segler, M. H. S. & Waller, M. P. Neural-Symbolic Machine Learning for Retrosynthesis and Reaction Prediction. *Chemistry* **23**, 5966-5971, doi:10.1002/chem.201605499 (2017).

9   Tetko, I. V., Karpov, P., Van Deursen, R. & Godin, G. State-of-the-art augmented NLP transformer models for direct and single-step retrosynthesis. *Nature Communications* **11**, 5575, doi:10.1038/s41467-020-19266-y (2020).

10  Sutskever, I., Vinyals, O. & Le, Q. V. Sequence to Sequence Learning with Neural Networks. arXiv:1409.3215 (2014). <https://ui.adsabs.harvard.edu/abs/2014arXiv1409.3215S>.

11  Cadeddu, A., Wylie, E. K., Jurczak, J., Wampler-Doty, M. & Grzybowski, B. A. Organic Chemistry as a Language and the Implications of Chemical Linguistics for Structural and Retrosynthetic Analyses. *Angewandte Chemie International Edition* **53**, 8108-8112, doi:10.1002/anie.201403708 (2014).

12  Jin, W., Coley, C. W., Barzilay, R. & Jaakkola, T. in *Proceedings of the 31st International Conference on Neural Information Processing Systems*    2604–2613 (Curran Associates Inc., Long Beach, California, USA, 2017).

13  Somnath, V. R., Bunne, C., Coley, C., Krause, A. & Barzilay, R. in *Advances in Neural Information Processing Systems.* (eds M. Ranzato *et al.*) 9405-9415 (Curran Associates, Inc.).

14  Weininger, D. SMILES, a chemical language and information system. 1. Introduction to methodology and encoding rules. *Journal of Chemical Information and Computer Sciences* **28**, 31-36, doi:10.1021/ci00057a005 (1988).

15  Krenn, M., Häse, F., Nigam, A., Friederich, P. & Aspuru-Guzik, A. Self-referencing embedded strings (SELFIES): A 100% robust molecular string representation. *Machine Learning: Science and Technology* **1**, 045024, doi:10.1088/2632-2153/aba947 (2020).

16  Liu, B. *et al.* Retrosynthetic Reaction Prediction Using Neural Sequence-to-Sequence Models. *ACS Central Science* **3**, 1103-1113, doi:10.1021/acscentsci.7b00303 (2017).

17  Hochreiter, S. & Schmidhuber, J. Long Short-term Memory. *Neural computation* **9**, 1735-1780, doi:10.1162/neco.1997.9.8.1735 (1997).

18  Vaswani, A. *et al.* Attention Is All You Need. arXiv:1706.03762 (2017). <https://ui.adsabs.harvard.edu/abs/2017arXiv170603762V>.

19  Karpov, P., Godin, G. & Tetko, I. V. in *Artificial Neural Networks and Machine Learning – ICANN 2019: Workshop and Special Sessions.* (eds Igor V. Tetko, Věra Kůrková, Pavel Karpov, & Fabian Theis) 817-830 (Springer International Publishing).

20      Irwin, R., Dimitriadis, S., He, J. & Bjerrum, E. J. Chemformer: a pre-trained transformer for computational chemistry. *Machine Learning: Science and Technology* **3** (2022).

21      Ucak, U. V., Ashyrmamatov, I., Ko, J. & Lee, J. Retrosynthetic reaction pathway prediction through neural machine translation of atomic environments. *Nature Communications* **13**, 1186, doi:10.1038/s41467-022-28857-w (2022).

22      Zheng, S., Jiahua, R., Zhang, Z., Xu, J. & Yang, Y. Predicting Retrosynthetic Reactions Using Self-Corrected Transformer Neural Networks. *Journal of chemical information and modeling* **60**, doi:10.1021/acs.jcim.9b00949 (2019).

23      Weisfeiler, B. & Leman, A. A reduction of a Graph to a Canonical Form and an Algebra Arising during this Reduction (in Russian). *Nauchno-Technicheskaya Informatsia* **9** (1968).

24      Shi, C., Xu, M., Guo, H., Zhang, M. & Tang, J. in *Proceedings of the 37th International Conference on Machine Learning*    Article 818 (JMLR.org, 2020).

25      Schlichtkrull, M. *et al.* in *The Semantic Web.* (eds Aldo Gangemi *et al.*) 593-607 (Springer International Publishing).

26      Yan, C. *et al.* in *Advances in Neural Information Processing Systems.* (eds H. Larochelle *et al.*) 11248-11258 (Curran Associates, Inc.).

27      Veličković, P. *et al.* in *International Conference on Learning Representations.*

28      Gilmer, J., Schoenholz, S. S., Riley, P. F., Vinyals, O. & Dahl, G. E. in *Proceedings of the 34th International Conference on Machine Learning - Volume 70*    1263–1272 (JMLR.org, Sydney, NSW, Australia, 2017).

29      Dai, H., Li, C., Coley, C. W., Dai, B. & Song, L. in *Proceedings of the 33rd International Conference on Neural Information Processing Systems*    Article 796 (Curran Associates Inc., 2019).

30      Chengxuan, Y. *et al.* in *Thirty-Fifth Conference on Neural Information Processing Systems*    (2021).

31      Halgren, T. A. Merck molecular force field. I. Basis, form, scope, parameterization, and performance of MMFF94. *Journal of Computational Chemistry* **17**, 490-519, doi:https://doi.org/10.1002/(SICI)1096-987X(199604)17:5/6<490::AID-JCC1>3.0.CO;2-P (1996).

32      Wang, Y., Wang, J., Cao, Z. & Barati Farimani, A. Molecular contrastive learning of representations via graph neural networks. *Nature Machine Intelligence* **4**, 279-287, doi:10.1038/s42256-022-00447-x (2022).

33      Stärk, H. *et al.* in *ICML.*

34      Khosla, P. *et al.* Supervised Contrastive Learning. arXiv:2004.11362 (2020). <https://ui.adsabs.harvard.edu/abs/2020arXiv200411362K>.

35      Schneider, N., Stiefl, N. & Landrum, G. A. What's What: The (Nearly) Definitive Guide to Reaction Role Assignment. *Journal of Chemical Information and Modeling* **56**, 2336-2346, doi:10.1021/acs.jcim.6b00564 (2016).

36      Coley, C. W., Rogers, L., Green, W. H. & Jensen, K. F. Computer-Assisted Retrosynthesis Based on Molecular Similarity. *ACS Central Science* **3**, 1237-1245, doi:10.1021/acscentsci.7b00355 (2017).

37      Segler, M. H. S. & Waller, M. P. Neural-Symbolic Machine Learning for Retrosynthesis and Reaction Prediction. *Chemistry – A European Journal* **23**, 5966-5971,

doi:https://doi.org/10.1002/chem.201605499 (2017).

38  Zheng, S., Rao, J., Zhang, Z., Xu, J. & Yang, Y. Predicting Retrosynthetic Reactions Using Self-Corrected Transformer Neural Networks. *Journal of Chemical Information and Modeling* **60**, 47-55, doi:10.1021/acs.jcim.9b00949 (2020).

39  Sun, R., Dai, H., Li, L., Kearnes, S. M. & Dai, B. in *NeurIPS*.

40  Mao, K. *et al.* Molecular graph enhanced transformer for retrosynthesis prediction. *Neurocomputing* **457**, 193-202, doi:https://doi.org/10.1016/j.neucom.2021.06.037 (2021).

41  Lin, K., Xu, Y., Pei, J. & Lai, L. Automatic retrosynthetic route planning using template-free models. *Chemical Science* **11**, 3355-3364, doi:10.1039/C9SC03666K (2020).

42  Kim, E., Lee, D., Kwon, Y., Park, M. S. & Choi, Y.-S. Valid, Plausible, and Diverse Retrosynthesis Using Tied Two-Way Transformers with Latent Variables. *Journal of Chemical Information and Modeling* **61**, 123-133, doi:10.1021/acs.jcim.0c01074 (2021).

43  Lin, Z., Yin, S., Shi, L., Zhou, W. & Zhang, Y. G2GT: Retrosynthesis Prediction with Graph to Graph Attention Neural Network and Self-Training. arXiv:2204.08608 (2022). <https://ui.adsabs.harvard.edu/abs/2022arXiv220408608L>.

44  Tu, Z. & Coley, C. W. Permutation invariant graph-to-sequence model for template-free retrosynthesis and reaction prediction. *Journal of chemical information and modeling* (2022).

45  Wang, X. *et al.* RetroPrime: A Diverse, Plausible and Transformer-based Method for Single-Step Retrosynthesis Predictions. *Chemical Engineering Journal* **420**, 129845, doi:10.1016/j.cej.2021.129845 (2021).

46  Seo, S. *et al.* in *AAAI*.

47  Mayer, C. D., Lorent, J. & Horgan, G. W. Exploratory analysis of multiple omics datasets using the adjusted RV coefficient. *Stat Appl Genet Mol Biol* **10**, Article 14, doi:10.2202/1544-6115.1540 (2011).

48  Khosla, P. *et al.* Supervised Contrastive Learning. *ArXiv* **abs/2004.11362** (2020).

49  Chen, B., Li, C., Dai, H. & Song, L. in *International Conference on Machine Learning (ICML).* (2020).